# Medium-Term Load Forecasting Using Support Vector Regression, Feature Selection, and Symbiotic Organism Search Optimization

Arghavan Zare-Noghabi; Morteza Shabanzadeh
Department of Power System Operation and Planning
Niroo Research Institute (NRI)
Tehran, Iran
a.zarenoghabi@mail.sbu.ac.ir; mshabanzadeh@nri.ac.ir

Hossein Sangrody
Electrical and Computer Engineering Department
State University of New York, Binghamton University
Binghamton, NY, 13902, USA
hsangro1@binghamton.edu

*Abstract*— An accurate load forecasting has always been one of the main indispensable parts in the operation and planning of power systems. Among different time horizons of forecasting, while short-term load forecasting (STLF) and long-term load forecasting (LTLF) have respectively got benefits of accurate predictors and probabilistic forecasting, medium-term load forecasting (MTLF) requires more attentions due to its vital role in power system operation and planning such as optimal scheduling of generation units, robust planning programs for customer service, and economic supply. In this study, a hybrid method, composed of Support Vector Regression (SVR) and Symbiotic Organism Search Optimization (SOSO) method, is proposed for MTLF. In the proposed forecasting model, SVR is the main part of the forecasting algorithm while SOSO is embedded into it to optimize the parameters of SVR. In addition, a minimum redundancy-maximum relevance feature selection algorithm is applied in the preprocessing of input data. The proposed method is tested on EUNITE competition dataset and compared with some previous works to demonstrate its high performance.

## I. INTRODUCTION

By considering influential parameters of load forecasting in different time horizons and uncertainty in the inputs of forecasting models, it can be deduced that the longer time horizon is, the more challenge the load forecasting will be to get accurate point estimation of future demand. In this regards, most of studies in load forecasting era have focused on very short-term and short- term load forecasting (VSTLF and STLF), in which the forecasting models get benefits of accurate predictors like accurate forecasted weather variables as well as influential lag values of historical load data [1]. Thus, the forecasting results in VSTLF and STLF have usually led to fairly accurate results. On the other hand, studies in long term load forecasting (LTLF) tend to implement probabilistic modelling instead of point estimation [2]. In this condition, medium-term load forecasting (MTLF) requires more attentions, considering its vital applications in the operation, control, and planning of power systems at generation, transmission, distribution, and marketing levels. Accurate MTLF is indispensable to have a better scheduling and planning programs in unit commitment, control of the system considering the unprecedented presence of distributed energy resources (DERs), hydro-thermal coordination, economic supply of different fuels, efficiency assessment, and the management of limited-energy [3, 4].

In literatures, the time horizon of MTLF are not considered the same. For instance, authors in [5] believe that the forecasting term of MTLF is 1 to 12 months ahead of time while, the load forecast within 3 months to 3 years is suggested by [6] as the medium-term time horizon. However, in another point of view, MTLF is carried out by [7] for a period between one to several years.

Considering the high performance of *Support Vector Regression* (SVR), several studies have used this method as the kernel of their forecasting models for different time horizons. In [8] , the authors have introduced SVR as one of most efficient machine learning methods in load forecasting. In [9], the electric load is predicted for 168 hours ahead using SVR in which significant temperature changes have been also taken into account. The reported result shows that serious changes in the temperature will definitely result in big error in the training of SVR using historical data. To overcome this problem, the authors have proposed a similar-day approach in case of temperature volatility. In [10], a general method based on SVR is used for STLF. Reference [11] has proposed a hybrid method consisting of SVR and *Ant Colony Optimization* (ACO) in which ACO is only used to process a large amount of data to omit redundant information. The authors of [12] have proposed a long-term load forecasting method using the combination of SVR and *Differential Evolution* (DE) algorithm. In this method, DE is used to find optimal values of SVR parameters. The capability of this method is compared with a pure SVR, regression, and back propagation neural network, and the simulation results show that the proposed model outperforms aforementioned methods.

Moreover, an SVR-based method is proposed in [13] where the optimization algorithm used for finding the optimal parameters of SVR is a combination of *Simulated Annealing Algorithm* and *Particle Swarm Optimization* (PSO) algorithm. Authors in [14] have also combined *Seasonal Auto Regressive Integrated Moving Average* (SARIMA) method, *Seasonal Exponential Smoothing* model, and weighted SVR and proposed a method that overcomes nonlinearity of data. In [15], chaotic PSO is used to find SVR parameters for long-term load forecasting which lead to outperformance of the proposed method to several other methods. A modified *Firefly Algorithm* with a better performance than PSO is used in [16] to determine SVR parameters. In [17], SVR is embedded into a neural network for the purpose of building a model for STLF.

In this paper, a high performance MTLF model based on *minimum redundancy maximum relevancy* (MRMR) feature selection algorithm, SVR, and a recently introduced optimization algorithm named *Symbiotic Organism Search* (SOS) [18] is proposed. The proposed methodology is applied for daily peak load forecasting of one month ahead. In this hybrid model, MRMR feature selection algorithm whose adequate performance is previously proven in [19, 20] are applied to choose those parameters that are the most relevant to target values while redundant features are omitted from input data set. The selected inputs are applied to the SVR model where some parameters of SVR should also be well-tuned. For this purpose, the high performance optimization algorithm of SOS is implemented.

The paper is organized as follows. In Section II, the main technical aspects of the proposed model, as well as the characteristics of its components are explained. In Section III, our method is tested on EUNITE data and the results are compared with some previous works. Conclusion and some challenges to be investigated in future are also discussed in Section IV.

## II. PROPOSED METHOD

### A. Support Vector Regression (SVR)

Support Vector Machine (SVM) was firstly introduced by *Vapnik* in 1995 [21]. While SVM is used for classification, SVR is applied for prediction with the same concepts as SVM. Consider training data $(x_i, y_i)$ which $x_i$ is input feature vector and $y_i$ is the output vector. If we consider $t_i$ as estimation vector of $y_i$, it can be show as follow:

$$t_i \approx y_i = w^t \phi(x_i) + b \qquad (1)$$

where, $\omega$ is the weight vector of each input and $b$ is the bias term. $\phi$ is a function that maps $x_i$ to a higher dimensional space. Due to the high ability of $\phi$ in nonlinear regression, it is one of kernel functions. In this paper, it is defined as a radial basis function. The model must have a limited error, therefore $\epsilon$ is considered as the maximum error that the model can tolerate. A loss function is needed to be defined as shown in Fig. (1) which are represented by (2) and (3):

$$L_\epsilon = \begin{cases} 0 & |t_i - y_i| \leq \epsilon \\ |t_i - y_i| - \epsilon & otherwise \end{cases} \qquad (2)$$

$$\xi_i = |t_i - y_i| - \epsilon \qquad (3)$$

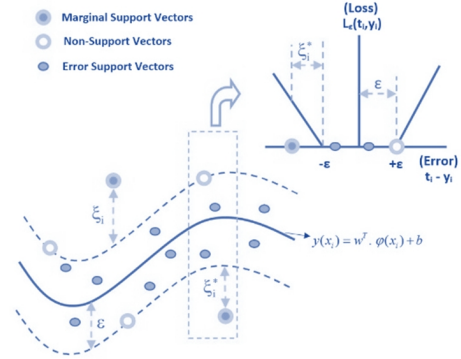

Figure 1. Conceptual model of SVR and its ε-tube loss function

If any one of data is outside of the $\epsilon$-tube function, a penalty cost defined by *C* should be considered. The optimization problem solved by SVR is as follows [22].

$$\min_{\omega, b, \xi, \xi^*} \frac{1}{2} \omega^T \omega + c \sum_{i=1}^{l} (\xi_i + \xi_i^*) \qquad (4)$$

$$\begin{cases} y_i - (\omega^T \phi(x_i) + b) \leq \epsilon + \xi_i^* \\ (\omega^T \phi(x_i) + b) - y_i \leq \epsilon + \xi_i \end{cases} \quad \xi_i, \xi_i^* \geq 0 \; i = 1, \dots, l \qquad (5)$$

where, $\xi_i^*$ is the upper limit of training error and $\xi_i$ is the lower limit of training error. $C$, $\epsilon$ and the parameters of kernel function are the ones that control regression quality. In order to avoid undesirable solutions using try-and-error approaches in tuning the proper values of these parameters, in our proposed method, they are determined by a novel optimization algorithm i.e., SOS.

### B. Maintaining the Integrity of the Specifications

The feature selection problem can be defined as finding a subset of features among a large number of feature sets that may characterize the target classification variables. There are many feature selection methods and the aim of these methods is to reduce dimension of problem while increasing accuracy of classification. *Minimum redundancy - maximum relevancy* (MRMR) feature selection method selects the maximum relevance features with minimum redundancy among introduced features. Using MRMR, a feature subset is selected in such a way that the statistical properties of a target classification variable are characterized at the best, subject to the constraints that these features are mutually as dissimilar to each other as possible, but marginally as similar to the classification variable as possible. This method actually consists of two separate methods of maximal relevancy and minimal redundancy. Maximal relevance feature selection selects the features with the highest relevance to the target class. Either correlation or mutual information may be the measure used to define dependency of variables. Minimum redundancy feature selection is also an algorithm that can be frequently used in a method to accurately identify characteristics of targets. The criterion combining these two aforementioned constraints is called MRMR. In fact, MRMR is useful to select features that are mutually far away from each other while at the same time having high association with the criterion.

*1) Maximal Relevancy*

Considering a feature set S with m features $\{x_i\}$ and $c$ as target, mutual information between feature set and the target can be shown as $I(x_i,c)$. The objective in maximal relevance, is to select features that individually have the largest mutual information $I(x_i, c)$ with the target class c, reflecting the largest dependency on the target class.

$$max\ D(S,c),\quad D = I(\{x_i, i = 1, \ldots, m\}, c) \quad (6)$$

$D(S,c)$ can be approximated with mean value of all mutual information values between $c$ and $x_i$ as below.

$$D(S,c) = \frac{1}{|S|} \sum_{x_i \in S} I(x_i, c) \quad (7)$$

*2) Minimal Redundancy*

The subsets that are identified by the maximum relevance are often contain sets which are relevant but redundant. Hence, using minimal redundancy helps to address this problem by removing those redundant subsets. The minimal redundancy condition can be written as below.

$$min\ R(S) = \frac{1}{|S|^2} \sum_{x_i, x_j \in S} I(x_i, x_j) \quad (8)$$

*3) Minimum Redundancy Maximum Relevancy*

The operator $ø(D, R)$ is defined as the following simple form to combine $D$ and $R$ while optimizing both of them simultaneously.

$$max\ ø(D, R),\quad ø = D - R \quad (9)$$

*C. Symbiotic Organism System Optimization Algorithm*

Symbiotic Organism System Optimization (SOSO) algorithm is a newly introduced algorithm by [18]. This algorithm is based on interactions of organisms in nature. Interactions between living organisms can be divided into three categories. The first category is *mutualism* in which two organisms get benefit of each other. The relationship between 'oxpeckers' and 'zebras'. Oxpecker is a bird that lives on the body of a zebra and feeds on the bugs and parasites from its body. The second category is *commensalism*. In this kind of relationship, one organism obtains food or other benefits without benefiting or harming the other one. An example of commensalism relationship is the connection between Pseudo scorpion and other insects. It is a kind of scorpion which is very tiny and it hitches rides on much larger insects from one place to another without harming them. In the third category, called *parasitism*, one organism builds a relationship with other organism to benefit itself while the other one is harmed, like parasites and human. The steps of this algorithm are shown in Fig. 2.

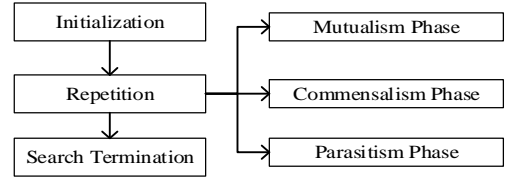

Figure 2. The main phases of the SOS algorithm

The initial population is called ecosystem and it consists of a group of organisms. For the purpose of searching the space, first, random values are generated for organisms between their upper and lower limits. Then, organism $i$ is randomly selected. Then, three main phases of SOS are performed respectively. Finally, the best value which corresponds to the minimum value of the fitness function is determined.

*1) Mutual phase*

In this phase, another organism named $j$ is also selected randomly. As shown in (10) and (11), each organism tries to improve its fitness value. In nature, irrespective of the profit obtained by the relationship of these two organisms, the amount of profits gained by an organism might not the same as the other one. To model this behavior, the parameter *BF* is defined in (10) and (11) as a profitability coefficient which is usually 1 and 2. Fitness values are obtained with these new values of organisms. If the last fitness value is better than the old one, it will be chosen as the best fitness value. Meanwhile, (12) represents the relationship characteristic between organism $x_i$ and $x_j$.

$$x_{inew} = x_i + rand(0,1) * (x_{best} - Mutual - vector * BF_1) \quad (10)$$

$$x_{jnew} = x_j + rand(0,1) * (x_{best} - Mutual - vector * BF_2) \quad (11)$$

$$Mutual\_vector = \frac{x_i + x_j}{2} \quad (12)$$

*2) Commensalism phase*

The next step is commensalism phase in which organism $j$ is selected randomly among other organisms. As shown in (13), organism $i$ benefits while the value of organism $j$ is constant. As previous state, the output of the problem is obtained according to the new position $i$, and if the result is better than the previous state, the existing position is updated.

$$x_{inew} = x_i + rand(-1,1) * (x_{best} - x_j) \quad (13)$$

*3) Parasitism phase*

Similar to previous phases, here, organism $j$ must also be selected randomly. Afterwards, the value of organism $i$ is multiplied in a random number and builds the parasite vector. Next, the fitness value of organism $j$ and parasite vector is obtained. If fitness value of parasite vector is better than organism $j$, the value of organism $j$ will change to parasite vector. Otherwise, organism $j$ will remain immune to parasite vector.

*D. Proposed Method*

In our proposed method, MRMR is initially used to find 10 most relevant past loads for each daily peak load amongst 60 historical loads. Some binary features are used in the feature set as well. These binary features are month index, weekday index, and national holiday index. These binary features along with selected past loads are training data set, and their corresponding load is target. Afterward, initialization step of the optimization algorithm starts.

Random vectors are created as SVR parameters and the ecosystem with the best fitness value is selected as initial set. The next step is mutualism phase, in which mutual vector and new candidate solution are calculated through (10) to (12). New candidate solutions are compared to the old ones and fitter organisms are selected as solutions for the next iteration. commensalism is the next step. In commensalism, an organism is selected randomly and the best organism in last step is modified using (13) and in case that the modified organism is fitter than the previous one, it is considered as the new best organism. The last step of search algorithm is parasitism. In this step, the same as other steps, one organism is chosen randomly and a parasite vector is created from this organism and if fitness of this organism is better than the best organism, best organism will be replaced. The steps explained above will be done for each organism in ecosystem and the whole procedure will be done to the maximum iteration which is set by the user.

In the flowchart shown in Fig. 3, the proposed method is depicted. As explained above, it is obvious that in the first step, a feature selection method is used to find the most relevant feature with minimum redundancy among a huge number of features. Afterwards, selected features will be used as input variables for SVM, and finally, the iterative process based on SOSO will begin to find the best SVM parameters.

### III. SIMULATION RESULTS

The proposed method is implemented on EUNITE data set [23]. The given data is electric peak-load data from January 1997 to January 1999. The aim is to predict maximum daily load of January 1999. To do so, the data set is divided into two sets of training set and test set. Training set consists of daily peak loads during the intervals of January to March and October to December in 1997 and 1998 while, the test set is daily peak load of January 1999.

Using feature selection method, 10 features related to past loads are selected i.e., 1, 2, 3, 4, 6, 7, 8, 14, 26 and 28 days before the forecasting day. In addition, some other features such as holidays, day type of a week, and month of a year have been taken into account in inputs by binary features. Each week is also divided into three types, namely, the first day of the week, weekdays, and weekends. Months are also determined by binary numbers. In this study, to evaluate the forecasting results, the most common error metric in load forecasting i.e. mean absolute percentage error (MAPE) is used. The results of load forecasting are demonstrated in Fig. 4 where the dashed line represent the predicted values of load and continuous line shows real load values. As seen, the real values of load are predicted satisfactorily by the proposed method. The error of the proposed method in load forecasting of 30 days ahead is 1.3904 %.

In order to compare the performance of the proposed method, some other methods are also simulated on the same data sets. Tables I shows the results of daily peak load forecasting which indicates the superiority of our proposed approach in compared to other methods. Moreover, for both teaching and learning purposes, the main load forecasting method with the kernel of SOS optimization algorithm is used again with and without feature selection method to reveal the influence of using this method in outputs. The results are shown in Table II. As seen, even without using feature selection method but using the same features, the SOS algorithm has a better performance than PSO which is known as one of the best global searching optimization algorithms.

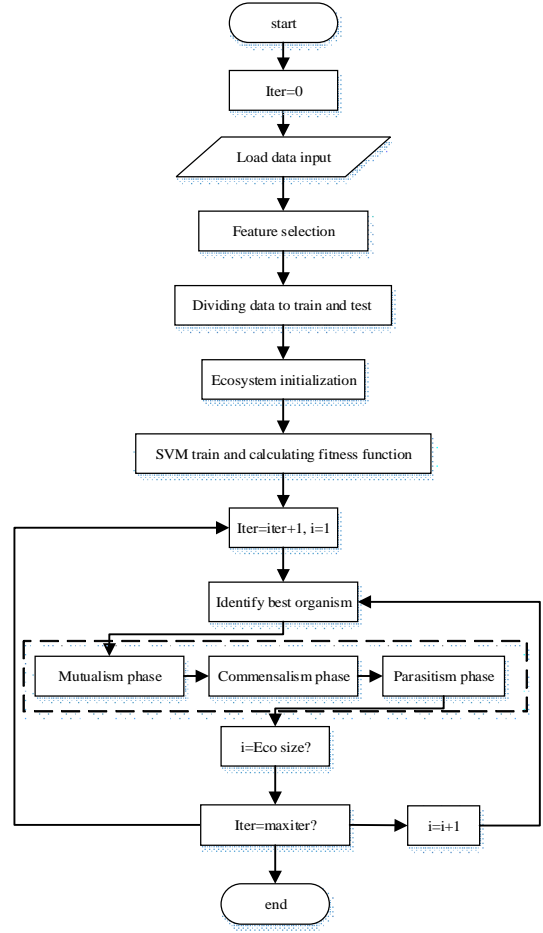

Figure 3. The flowchart of the proposed method

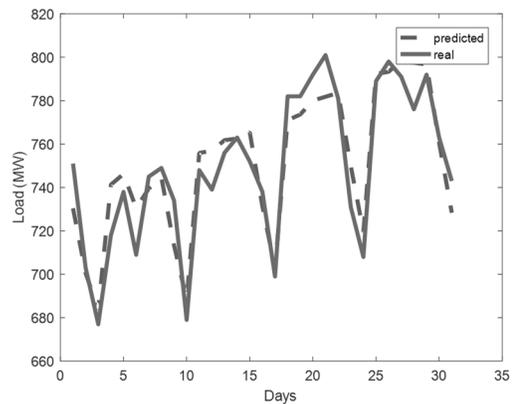

Figure 4. The load forecasts using the proposed method

TABLE I. Forecasting accuracy of the proposed MTLF tool

| Methods | MAPE (%) |
| --- | --- |
| Theta | 5.77 |
| Trigonometry` smoothing | 4.77 |
| Exponential smoothing | 4.67 |
| ETS | 4.64 |
| ARIMA | 3.77 |
| MLP neural network | 1.89 |
| Our proposed method | 1.39 |

TABLE II. Comparing the results between two powerful optimization kernels and with and without using feature selection algorithm

| Methods | MAPE (%) |
| --- | --- |
| PSO+ user selected features (*try-and-error*) | 1.5316 |
| PSO+ MRMR | 1.4305 |
| SOS+ user selected features (*try-and-error*) | 1.4171 |
| SOS+ MRMR | 1.3904 |

IV. CONCOLUSION

In this paper, a new hybrid method was proposed for MTLF. SOS was suggested as the optimization kernel to find optimal parameters of SVR while MRMR feature selection method was also used to find the most relevant features with minimum redundancy to avoid non-relevant and extra features. As illustrated in the case studies, in compared with PSO, SOS has a better performance in finding optimal parameters of SVR. Furthermore, feature selection techniques can help users to find most relevant features instead of try-and-error approaches and based on using a large number of features without any information about the most relevant ones. Another advantage of using feature selection methods is reducing the dimension of the main problem which makes it easier to be solved. Further research will address this issue for augmentation of SOS and embedding more efficient feature selection methods to improve the performance of our proposed method. Also, the number of most relevant features, with minimum redundancy can be optimized using optimization algorithms which constitute ground for future research work.

ACKNOWLEDGMENT

This work was supported by *Niroo Research Institute* (NRI) under Contract No. PONPN06. The authors would like to thank *Mrs. Zahra Madihi-Bidgoli* and *Professor Ali Keyhani* for their valuable comments and feedbacks on this research project, and the referees for their valuable comments and suggestions.

REFERENCES

[1] H. Sangrody, M. Sarailoo, N. Zhou, A. Shokrollahi, and E. Foruzan, "On the performance of forecasting models in the presence of input uncertainty," in *Power Symposium (NAPS), 2017 North American*, 2017, pp. 1-6.
[2] H. Sangrody, N. Zhou, and X. Qiao, "Probabilistic models for daily peak loads at distribution feeders," in *Power & Energy Society General Meeting, 2017 IEEE*, 2017, pp. 1-5.
[3] N. Amjady and F. Keynia, "Mid-term load forecasting of power systems by a new prediction method," *Energy Conversion and Management,* vol. 49, pp. 2678-2687, 2008.
[4] M. Ghiassi, D. K. Zimbra, and H. Saidane, "Medium term system load forecasting with a dynamic artificial neural network model," *Electric Power Systems Research,* vol. 76, pp. 302-316, 2006.
[5] N. Abu-Shikhah, F. Elkarmi, and O. M. Aloquili, "Medium-term electric load forecasting using multivariable linear and non-linear regression," *Smart Grid and Renewable Energy,* vol. 2, p. 126, 2011.
[6] P. Bunnoon, "Mid-Term Load Forecasting Based on Neural NetworkAlgorithm: a Comparison of Models," *International Journal of Computer and Electrical Engineering,* vol. 3, p. 600, 2011.
[7] G. J. Tsekouras, N. D. Hatziargyriou, and E. N. Dialynas, "An optimized adaptive neural network for annual midterm energy forecasting," *IEEE Transactions on Power Systems,* vol. 21, pp. 385-391, 2006.
[8] H. Sangrody, N. Zhou, S. Tutun, B. Khorramdel, M. Motalleb, and M. Sarailoo, "Long term forecasting using machine learning methods," in *Power and Energy Conference at Illinois (PECI), 2018 IEEE*, 2018, pp. 1-5.
[9] A. Selakov, D. Cvijetinović, L. Milović, S. Mellon, and D. Bekut, "Hybrid PSO–SVM method for short-term load forecasting during periods with significant temperature variations in city of Burbank," *Applied Soft Computing,* vol. 16, pp. 80-88, 2014.
[10] E. Ceperic, V. Ceperic, and A. Baric, "A strategy for short-term load forecasting by support vector regression machines," *IEEE Transactions on Power Systems,* vol. 28, pp. 4356-4364, 2013.
[11] D. Niu, Y. Wang, and D. D. Wu, "Power load forecasting using support vector machine and ant colony optimization," *Expert Systems with Applications,* vol. 37, pp. 2531-2539, 2010.
[12] J. Wang, L. Li, D. Niu, and Z. Tan, "An annual load forecasting model based on support vector regression with differential evolution algorithm," *Applied Energy,* vol. 94, pp. 65-70, 2012.
[13] J. Wang, Y. Zhou, and X. Chen, "Electricity load forecasting based on support vector machines and simulated annealing particle swarm optimization algorithm," in *Automation and Logistics, 2007 IEEE International Conference on*, 2007, pp. 2836-2841.
[14] J. Wang, S. Zhu, W. Zhang, and H. Lu, "Combined modeling for electric load forecasting with adaptive particle swarm optimization," *Energy,* vol. 35, pp. 1671-1678, 2010.
[15] W.-C. Hong, "Chaotic particle swarm optimization algorithm in a support vector regression electric load forecasting model," *Energy Conversion and Management,* vol. 50, pp. 105-117, 2009.
[16] A. Kavousi-Fard, H. Samet, and F. Marzbani, "A new hybrid modified firefly algorithm and support vector regression model for accurate short term load forecasting," *Expert systems with applications,* vol. 41, pp. 6047-6056, 2014.
[17] C.-N. Ko and C.-M. Lee, "Short-term load forecasting using SVR (support vector regression)-based radial basis function neural network with dual extended Kalman filter," *Energy,* vol. 49, pp. 413-422, 2013.
[18] M.-Y. Cheng and D. Prayogo, "Symbiotic organisms search: a new metaheuristic optimization algorithm," *Computers & Structures,* vol. 139, pp. 98-112, 2014.
[19] X. Jin, E. W. Ma, L. L. Cheng, and M. Pecht, "Health monitoring of cooling fans based on Mahalanobis distance with mRMR feature selection," *IEEE Transactions on Instrumentation and Measurement,* vol. 61, pp. 2222-2229, 2012.
[20] G. Chandrashekar and F. Sahin, "A survey on feature selection methods," *Computers & Electrical Engineering,* vol. 40, pp. 16-28, 2014.
[21] A. J. Smola and B. Schölkopf, "A tutorial on support vector regression," *Statistics and computing,* vol. 14, pp. 199-222, 2004.
[22] S. Akhlaghi, H. Sangrody, M. Sarailoo, and M. Rezaeiahari, "Efficient operation of residential solar panels with determination of the optimal tilt angle and optimal intervals based on forecasting model," *IET Renewable Power Generation,* vol. 11, pp. 1261-1267, 2017.
[23] G. Dudek, "Short-Term load forecasting using random forests," in *Intelligent Systems' 2014*, ed: Springer, 2015, pp. 821-828.